%% file: iclr2026_conference.tex
\documentclass{article} 
\usepackage{iclr2026_conference,times}
\usepackage{xspace}

\input{math_commands.tex}

\usepackage{hyperref}
\usepackage{url}
\usepackage{graphicx}
\usepackage{caption}
\usepackage{pifont}         
\usepackage{wrapfig}
\usepackage{float}
\usepackage{makecell}
\usepackage{enumitem}
\usepackage{graphicx}
\usepackage{booktabs}
\usepackage{amsmath} 
\usepackage{multicol}
\usepackage{multirow}
\usepackage{threeparttable}
\usepackage{wrapfig}

\usepackage{xcolor}

\newcommand{\eg}{\textit{e.g.,}\xspace}
\newcommand{\cmark}{\ding{51}} 
\newcommand{\xmark}{\ding{55}} 

\definecolor{green}{HTML}{9AFF99}
\definecolor{blue}{HTML}{96FFFB}
\definecolor{red}{HTML}{FFCCC9}

\title{NEBULA: Do We Evaluate Vision-Language-Action Agents Correctly?}

\author{
Jierui Peng\thanks{Equal contribution.},
~Yanyan Zhang\footnotemark[1],
~Yicheng Duan\footnotemark[1],
~Tuo Liang,
~Vipin Chaudhary,
~Yu Yin\thanks{Corresponding Author} \\
Department of Computer \& Data Sciences \\ Case Western Reserve University \\
\texttt{\{jierui.peng,yanyan.zhang,yicheng.duan,tuo.liang,vipin,yu.yin\}} \\
\texttt{@case.edu} \\
Homepage: \url{https://vulab-ai.github.io/NEBULA-Alpha/}
}

\iclrfinalcopy 
\begin{document}

\maketitle


\input{sections/0-abstract}


\input{sections/1-introduction}


\input{sections/2-related_works}


\input{sections/3-methods}


\input{sections/4-experiment}


\input{sections/5-discussion}


\input{sections/6-conclusion}

\newpage
\bibliography{iclr2026_conference}
\bibliographystyle{iclr2026_conference}

\newpage
\appendix
\section{Appendix}

\input{sections/appendix_tasks}

\end{document}

%% file: math_commands.tex
\usepackage{amsmath,amsfonts,bm}

\def\eqref#1{equation~\ref{#1}}

\def\1{\bm{1}}

\DeclareMathAlphabet{\mathsfit}{\encodingdefault}{\sfdefault}{m}{sl}
\SetMathAlphabet{\mathsfit}{bold}{\encodingdefault}{\sfdefault}{bx}{n}



%% file: sections/0-abstract.tex

\begin{abstract}

The evaluation of Vision-Language-Action (VLA) agents is hindered by the coarse, end-task success metric that fails to provide precise skill diagnosis or measure robustness to real-world perturbations. 
This challenge is exacerbated by a fragmented data landscape that impedes reproducible research and the development of generalist models. 
To address these limitations, we introduce \textbf{NEBULA}, a unified ecosystem for single-arm manipulation that enables diagnostic and reproducible evaluation. NEBULA features a novel dual-axis evaluation protocol that combines fine-grained \textit{capability tests} for precise skill diagnosis with systematic \textit{stress tests} that measure robustness.
A standardized API and a large-scale, aggregated dataset are provided to reduce fragmentation and support cross-dataset training and fair comparison.
Using NEBULA, we demonstrate that top-performing VLAs struggle with key capabilities such as spatial reasoning and dynamic adaptation, which are consistently obscured by conventional end-task success metrics. By measuring both what an agent can do and when it does so reliably, NEBULA provides a practical foundation for robust, general-purpose embodied agents.
\end{abstract}

%% file: sections/1-introduction.tex
\vspace{-0.5em}
\section{Introduction}
Vision–Language–Action (VLA) agents are advancing rapidly, spanning language-conditioned planners, generalist multi-modal agents, and prompt-conditioned manipulation policies~\citep{brohan2023can, zitkovich2023rt, jiang2022vima}. Yet a basic question remains: \textit{are we evaluating what actually matters?} 
Most benchmarks tend to prioritize end-task success, a coarse metric that neither reveals which subskills are engaged nor localizes error sources.
For example, a failure on ``pick-and-place'' may arise from language grounding, 3D perception, spatial planning, or control. However, a single success rate cannot identify the failing component. Without capability-resolved, diagnostic evaluation, we cannot measure per-skill capability and expose where and why agents fail.

Even with precise skill diagnosis, current evaluation overlooks a second deployment-critical dimension: reliability. Passing a test at a single operating point does not imply robustness, nor does it reflect key properties needed for deployment (\eg latency, stability, robustness). Small, realistic shifts in conditions (\eg lighting, textures, phrasing, dynamics, sensor noise) can flip outcomes, while aggregate success rates often hide variability across settings and mask abrupt breakdowns (`failure cliffs'). Because real-world conditions continually shift along these dimensions, stress tests are needed to characterize reliability boundaries and disentangle competence from robustness.


Meanwhile, this dual challenge of diagnostic and robust evaluation is compounded by a severely fragmented data landscape. Datasets like ManiSkill~\citep{mu2021maniskill}, LeRobot~\citep{cadene2024lerobot}, and BEHAVIOR-1k~\citep{li2023behavior} differ drastically in format, task representation, and embodiment. Even efforts like Open-X~\citep{open_x_embodiment_rt_x_2023}, which propose shared interfaces, fall short in defining what capabilities are tested or how to compare them. 
This fragmentation forces researchers to reimplement pipelines, prevents fair head-to-head comparisons, and limits large-scale generalization studies, which slows progress toward unified embodied intelligence. As a result, the field lacks a unified ecosystem that can simultaneously diagnose agent capabilities, stress-test their robustness, and unify disparate data sources for reproducible, scalable research.


To address these challenges, we introduce \textbf{NEBULA}, an integrated ecosystem designed to shift the focus of embodied AI research from simple task completion to true capability mastery. The ecosystem is built on two core pillars. 
The first is a \textbf{diagnostic evaluation framework} that transforms coarse success rates into an interpretable, multi-faceted signal. It directly confronts the issues of disentangled capability evaluation and reliability by combining: (1) \textbf{Capability Tests}, which isolate specific skills like spatial reasoning and grasp synthesis to pinpoint precise reasons for failure, and (2) \textbf{Stress Tests}, which systematically vary environmental conditions to map an agent’s robustness and identify hidden failure cliffs. This dual-axis approach provides a holistic view of an agent's competence, revealing not only what it can do but also the conditions under which it can be trusted.


\begin{figure}[t!]
    \centering
    \includegraphics[width=\textwidth]{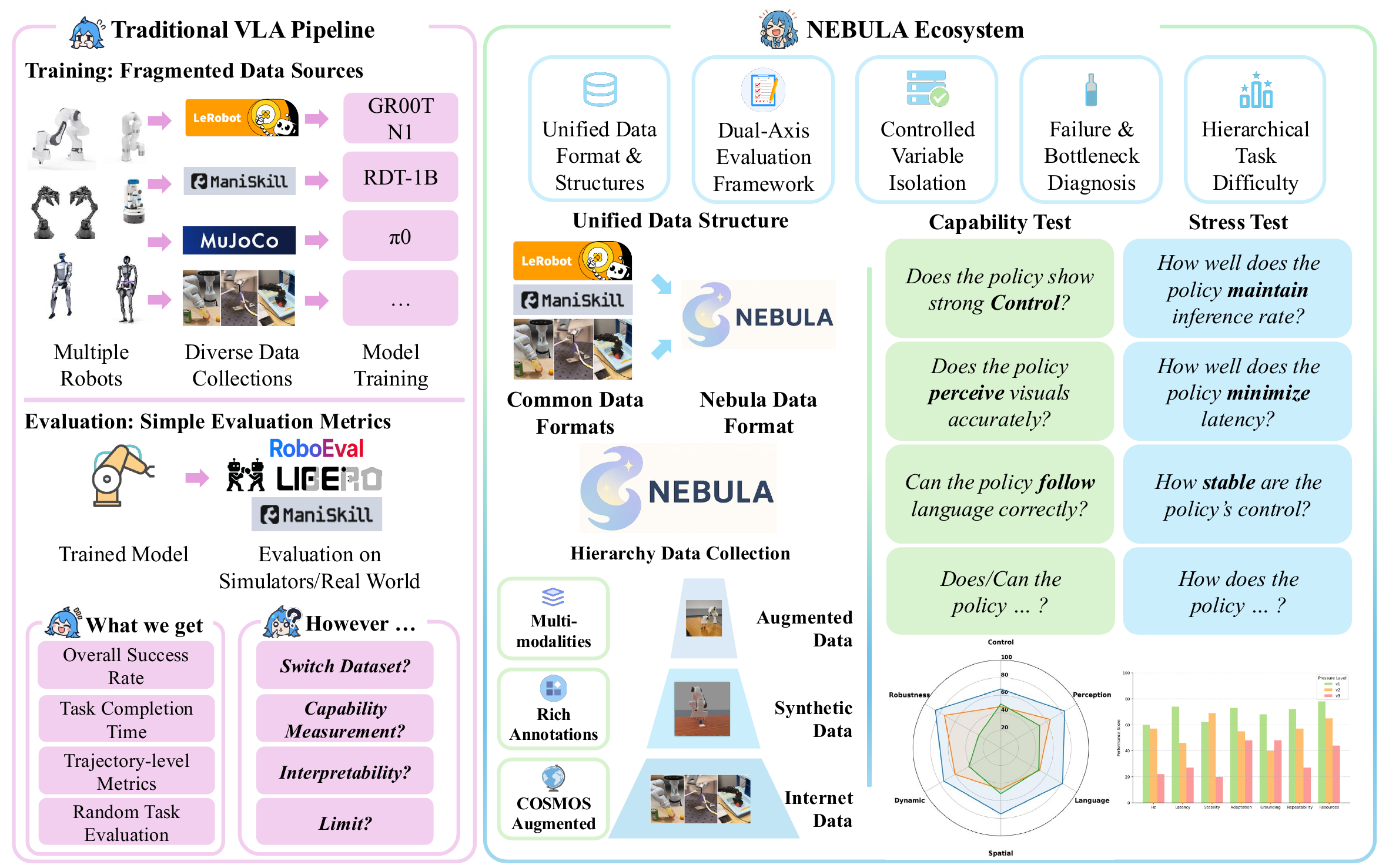}
    \caption{\textbf{NEBULA Ecosystem} unifies fragmented VLA datasets and APIs for cross-dataset training and benchmarking. It introduces a dual-axis evaluation (capability and stress testing) with controlled variable isolation for skill-specific diagnosis. With hierarchical task difficulty, multi-modal annotations, and visual performance summaries, NEBULA converts success rate into a diagnostic signal, exposing failure modes and reliability limits.}
    \vspace{-5mm}
    \label{fig:overall}
\end{figure}

Complementing its evaluation suite, NEBULA’s second pillar tackles the critical issue of \textbf{data and tooling fragmentation}. We provide a standardized API and data format that unifies disparate benchmarks, including ManiSkill, LeRobot, and others, eliminating the need for engineering work for each new dataset. We also provide a large-scale, aggregated dataset that integrates existing real-world demonstrations, simulator-generated trajectories, and world-model–augmented data. By providing the infrastructure for both unified training and reproducible evaluation, NEBULA empowers researchers to build more generalizable agents and conduct fair, large-scale comparisons that accelerate scientific progress.
In summary, the key contributions of our paper include:
\begin{itemize}[leftmargin=*,noitemsep,nolistsep]
    \item We introduce \textbf{NEBULA}, a unified VLA ecosystem that provides a standardized API and a large-scale, aggregated dataset to facilitate reproducible, cross-dataset training and benchmarking.
    \smallskip
    \item We propose a novel \textbf{dual-axis evaluation protocol} that combines fine-grained \textit{capability tests} for precise skill diagnosis with systematic \textit{stress tests} to measure an agent's robustness against real-world perturbations.
    \smallskip
    \item We present an \textbf{in-depth benchmarking study} of current VLAs, revealing critical failure modes (\eg spatial reasoning) that are typically obscured by the traditional success rate metric.
    \smallskip
\end{itemize}

%% file: sections/2-related_works.tex

\section{Related Works}
\subsection{Single-Arm Manipulation Benchmarks \& Simulators}

The landscape of robotic manipulation evaluation has expanded significantly in recent years, yet fundamental questions about what and how we measure remain unresolved. Existing efforts cluster into three threads: 
(i) \textbf{Single-arm tabletop benchmarks}, such as RLBench~\citep{james2020rlbench}, BulletArm~\citep{wang2022bulletarm}, ManiSkill2~\citep{gu2023maniskill2}, and ManiSkill3~\citep{tao2024maniskill3}, provide diverse task libraries, multimodal observations, and extensions toward bimanual, language-conditioned manipulation. 
(ii) \textbf{Long-horizon benchmarks}, such as BEHAVIOR-1K~\citep{li2023behavior}, Meta-World~\citep{yu2020meta}, ALFRED~\citep{shridhar2020alfred}, FurnitureBench~\citep{heo2023furniturebench}, Franka Kitchen~\citep{gupta2019relay}, LIBERO~\citep{liu2023libero}, CALVIN~\citep{mees2022calvin}, VLABench~\citep{zhang2024vlabench}, and MIKASA-Robo~\citep{cherepanov2025mikasa}, highlight multi-skill acquisition, temporal reasoning across extended tasks, and memory-centric challenges under partial observability.
(ii) \textbf{Realism-focused platforms}, such as SIMPLER~\citep{li2024evaluating}, Habitat~\citep{habitat19iccv}, SAPIEN~\citep{Xiang_2020_SAPIEN}, THE COLOSSEUM~\citep{pumacay2024colosseum}, and Genesis~\citep{Genesis}, advance physics fidelity and enable evaluation under controlled perturbations and language-conditioned tasks. 

Despite these advances, evaluation in most benchmarks still relies heavily on task-level success rate. While useful for model comparison and easy to compute, these metrics have limited diagnostic value: they indicate neither which abilities failed nor why. Our framework addresses this gap through a dual-axis evaluation that disentangles task requirements from performance quality, enabling structured and interpretable diagnosis.

\subsection{Evaluation Protocols \& Metrics}

Separating sources of failure is essential for evaluating VLA models in robotic manipulation, particularly as tasks grow more complex and as the demand for stronger generalization increases. THE COLOSSEUM~\citep{pumacay2024colosseum} systematically perturbs tasks along controlled axes and reports robustness degradation. VLABench~\citep{zhang2024vlabench} divides evaluation into six high-level capability dimensions to assess models more explicitly. RAMP (Robotic Assembly Manipulation and Planning)~\citep{collins2023ramp} introduces long-horizon assembly scenarios that challenge reasoning, diagnostics, and fault recovery in addition to pure control and perception. Meanwhile, Robot Policy Evaluation for Sim-to-Real Transfer~\citep{yang2025robot} proposes benchmarking strategies that gradually increase task complexity and introduce scenario perturbations to assess robustness and alignment between simulation and real-world performance. Also, Recent surveys on VLA models emphasize the need for evaluation across the full perception–language–control pipeline, combining task success with metrics for generalization, robustness, and instruction understanding~\citep{ma2024survey, shao2025large, sapkota2025vision}. Our work builds on the idea of evaluating intelligence across multiple dimensions, and further enforces protocols that disentangle task specifications from execution performance via controlled variation and progressively increasing difficulty.

%% file: sections/3-methods.tex
\vspace{-0.5em}
\section{Nebula Ecosystem}
NEBULA is a unified and comprehensive ecosystem built to overcome critical limitations in existing Embodied AI pipelines. While traditional systems often reduce evaluation to coarse metrics like task success or runtime, NEBULA broadens the scope to answer a deeper question: \textit{how and why does an agent succeed or fail?} As shown in Figure \ref{fig:overall}, NEBULA provides a structured, modular framework that includes 1) a standardized data layer with a unified format and APIs to enable cross-task training and reuse; 2) a dual-axis evaluation protocol for disentangling functional capabilities from real-time robustness; and 3) rich diagnostic outputs to support interpretable, skill-specific performance analysis. This section introduces NEBULA's core design. Section~\ref{subsec:data_specification} details our data collection protocol and unified API design, while Section~\ref{subsec:dual_evaluation} outlines NEBULA’s evaluation framework and the design of its capability and stress test tasks.


\subsection{Data \& API Specification}
\label{subsec:data_specification}

\noindent\textbf{Data Collection \& Annotation.}
To ensure consistency and reproducibility, NEBULA collects all training and evaluation data using a customized simulation platform built upon the SAPIEN~\citep{Xiang_2020_SAPIEN} engine and the ManiSkill3~\citep{tao2024maniskill3} framework. For each manipulation episode, we record a temporally ordered sequence of multimodal observations $\mathcal{O}_t$, system states $\mathcal{S}_t$, actions $\mathcal{A}_t$, and binary success labels $\mathcal{SU}_t \in {0, 1}$ at each timestep $t$. The observations $\mathcal{O}_t$ include RGB, depth, and segmentation images from six fixed-viewpoint cameras, as well as proprioceptive inputs such as joint positions $q_t$ and velocities $\Tilde{q}_t$. Each episode is annotated with a natural language task instruction, manually written to reflect the intended goal. These instructions serve as the conditioning input for language-conditioned policies and allow precise alignment between episodes and their semantic objectives. NEBULA offers two dataset variants, Alpha and Beta, designed to balance completeness and usability. For data collection, the Alpha version of the dataset is entirely generated using expert trajectories produced via motion planning~\citep{lavalle2006planning}. In contrast, the Beta version combines motion planning with human teleoperation: for selected hard tasks, expert demonstrations are collected manually to capture more diverse and realistic behaviors.

\noindent\textbf{API \& Modulated Assets.} To ensure consistency and ease of use across heterogeneous data sources, we introduce a unified data schema that consolidates fields found in modern embodied datasets. This schema standardizes the representation of observations, actions, environment states, and task metadata under a common structure, enabling plug-and-play compatibility with a wide range of learning algorithms. We provide a PyTorch API that abstracts away the low-level data loading and indexing details, exposing a clean, task-agnostic interface for pipeline. For researchers working in the TensorFlow ecosystem, we additionally provide lightweight TF-compatible adapters. To further reduce integration overhead, we include model-specific adapters for several widely used architectures, allowing for immediate benchmarking on NEBULA data with minimal code changes. Please refer to Appendix~\ref{appendix:unified_data_platform} for detailed information.


\begin{wraptable}{r}{0.48\linewidth}
\vspace{-4mm}
\caption{Dataset statistics of NEBULA-Alpha across five task families, excluding Robustness.}
\centering
\Large
\setlength{\tabcolsep}{6pt} 
\renewcommand{\arraystretch}{1.0} 
\small
\vspace{-1em}
\resizebox{\linewidth}{!}{ 
\begin{tabular}{@{}cccc@{}}
\toprule
\multirow{2}{*}{\begin{tabular}[c]{@{}c@{}}Task\\ Families\end{tabular}} & \multicolumn{3}{c}{Alpha}        \\ \cmidrule(l){2-4} 
                                                                         & Videos  & Descriptions & Traj    \\ \midrule
Control                                                                  & 54,000  & 9            & 36,000  \\
Perception                                                               & 54,000  & 9,000        & 36,000  \\
Language                                                                 & 48,000  & 24,000       & 96,000  \\
Dynamic                                                                  & 36,000  & 6            & 24,000  \\
Spatial                                                                  & 30,000  & 5,000        & 24,000  \\
Robust                                                                   & N/A     & N/A          & N/A     \\ \midrule
Total                                                                    & 222,000 & 38,015       & 216,000 \\ \bottomrule
\end{tabular}
} 
\label{table:data_statistics}
\vspace{-2mm}
\end{wraptable}

\noindent\textbf{Dataset Statistics.} 
NEBULA offers two dataset variants—Alpha and Beta—for both full-scale evaluation and lightweight experimentation. As shown in Table~\ref{table:data_statistics}, Alpha includes over 54,000 expert demonstrations across five capability families, while Beta is a compact version (~10\% per task) designed for rapid development and ablation. Some high-difficulty Beta tasks use human teleoperation to introduce realistic variations. Both datasets provide multimodal inputs (videos, language, trajectories) in PyTorch and TFRecord formats with adapter support. The Robustness and Generalization family is reserved for evaluation only and excluded from both training sets to prevent overfitting and ensure fair comparison under distribution shift.

\definecolor{red}{RGB}{255,0,0}
\definecolor{green}{RGB}{0,255,0}
\definecolor{blue}{RGB}{0,0,255}
\begin{figure}[!t]
    \centering
    \includegraphics[width=\textwidth]{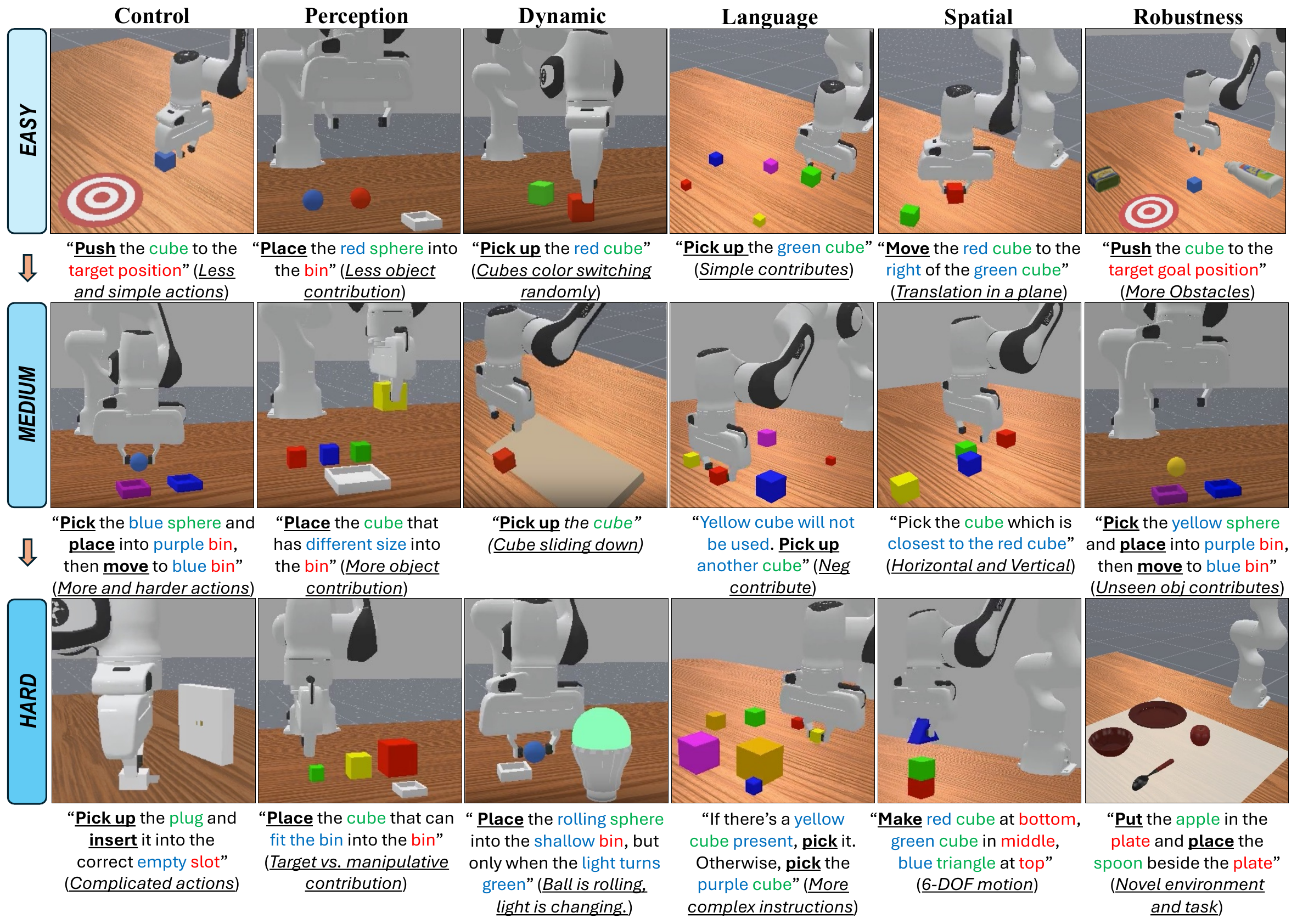}
    \caption{\textbf{Examples of NEBULA Capability Test task} across six core capabilities (Control, Perception, Dynamic Adaptation, Language, Spatial Reasoning, and Robustness) organized into three difficulty levels. Tasks isolate specific skills with controlled complexity. \textcolor{green}{Green} marks objects, \textcolor{red}{red} marks targets, and \textcolor{blue}{blue} indicates contextual cues. \textbf{\underline{Bold underlined}} text shows actions; \textit{\underline{italic underlined}} text gives clarifications.}
    \label{fig:capability_examples}
\end{figure}


\subsection{Dual-Axis Evaluation Framework}
\label{subsec:dual_evaluation}
NEBULA introduces a dual-axis evaluation framework to enable structured, interpretable, and diagnostic assessment of embodied AI systems. This framework decouples the evaluation into two dimensions: \textbf{Capability} and \textbf{Stress Tests}, each isolating a distinct facet of system performance. The Capability axis evaluates \emph{what the agent can do} under nominal conditions. The Stress axis probes \emph{how well the agent operates} under varying levels of real-time or robustness-related pressure.

\subsubsection{Capability Test Tasks}
NEBULA’s Capability Tests isolate six core embodied skills or capabilities through a suite of procedurally generated tasks. Our evaluation methodology is built on two key principles:
\textit{(i) Controlled-Variable Isolation}: Each task is designed to vary a single capability dimension while holding others constant, ensuring that performance changes can be unambiguously attributed to the skill being tested. For example, perception tasks minimize control complexity, while control tasks use fixed visual scenes.
\textit{(ii) Systematic Difficulty Scaling}: Within each family, tasks are generated from parameterized templates into three tiers (Easy, Medium, Hard), allowing for a fine-grained analysis of an agent's limits.
This modular structure enables reproducible, fine-grained evaluation of the following capabilities (see Fig.~\ref{fig:capability_examples} and Appendix~\ref{appendix:capabilty}):


\begin{enumerate}[label=(\arabic*), leftmargin=*, noitemsep, nolistsep]
    \item \textbf{\textit{Control}:} The Control task family isolates low-level manipulation by fixing non-control factors, with tasks progressing from simple actions (Easy) to precise, multi-step sequences (Hard).
    
    \smallskip
    \item \textbf{\textit{Perception}:} The Perception task family isolates visual recognition by minimizing control demands, with difficulty scaling from clear distinctions to subtle differences and cluttered scenes.
    
    \smallskip
    \item \textbf{\textit{Language}:} The Language task family tests instruction understanding, from basic grounding to reasoning and conditionals, with fixed scenes to isolate linguistic skills.
    
    \smallskip
    \item \textbf{\textit{Dynamic Adaptation}:} This task family evaluates how well an agent adapts to dynamic changes, from object attribute switching (Easy) to predictable moving (Medium) and unpredictable real-time events (Hard), testing reactivity and robustness.
    
    \smallskip
    \item \textbf{\textit{Spatial Reasoning}:} The Spatial task family tests spatial reasoning from 2D placement to 6-DoF planning, with difficulty scaling from planar to full 3D geometric understanding.
    
    \smallskip
    \item \textbf{\textit{Robustness/Generalization}:} The Robustness task family assesses generalization under distribution shifts, from distractors to unseen attributes to novel scenes.

\end{enumerate}







\subsubsection{Stress Test Tasks}
To complement capability-based evaluations, NEBULA introduces a suite of Stress Tests (\emph{Inference Frequency}, \emph{Latency}, \emph{Stability Score}, and \emph{Adaptation}) that isolate and quantify system performance under targeted operational constraints. Each test is a single-indicator probe. These tests avoid confounding variables and support controlled ablation studies by being independently applied. Each is instantiated at three calibrated pressure levels ($v_1$–$v_3$), defined by measurable parameters normalized to baseline conditions. This structure enables detailed stress-response profiling and fair comparisons across systems, helping identify bottlenecks and guide robustness optimization for real-world deployment. Full test definitions appear in Appendix~\ref{appendix:stress}.

\begin{enumerate}[label=(\arabic*), leftmargin=*, noitemsep, nolistsep]

    \item \textbf{\textit{Inference Frequency}:} This test measures action rate to assess real-time responsiveness, with increasing motion complexity exposing inference speed limits.

    \smallskip
    
    \item \textbf{\textit{Latency}:} This measures the delay from perception to action. Three tiers introduce increasing scene dynamics to responsiveness. Low latency is essential for precise, time-sensitive manipulation.

    \item \textbf{\textit{Stability Score}:} Stability quantifies action smoothness by measuring action variation between consecutive timesteps given an action sequence $\{a_{0}, a_{1}, ..., a_{t}\}$:
    \begin{equation}
    \vspace{-0.4em}
        \text{Stability} = \exp\left(-\frac{1}{T-1}\sum_{t=1}^{T}||\mathbf{a}_t - \mathbf{a}_{t-1}||_2\right)  
    \vspace{-0.1em}
    \end{equation}
    where $||\;||_2$ represents the $L_2$ nortm and higher scores ($\in [0,1]$) indicate smoother trajectories. Tests progress from coarse force control ($v_1$) to precise, contact-rich manipulation ($v_3$), revealing whether policies produce stable outputs suitable for deployment.

    \smallskip

    \item \textbf{\textit{Adaptability}:} Adaptability tests how well agents adjust to changing goals, from target shifts to instruction switches and rapid re-planning, revealing robustness under dynamic conditions.
\end{enumerate}





%% file: sections/4-experiment.tex

\begin{figure}[!t]
    \centering
    \includegraphics[width=\textwidth]{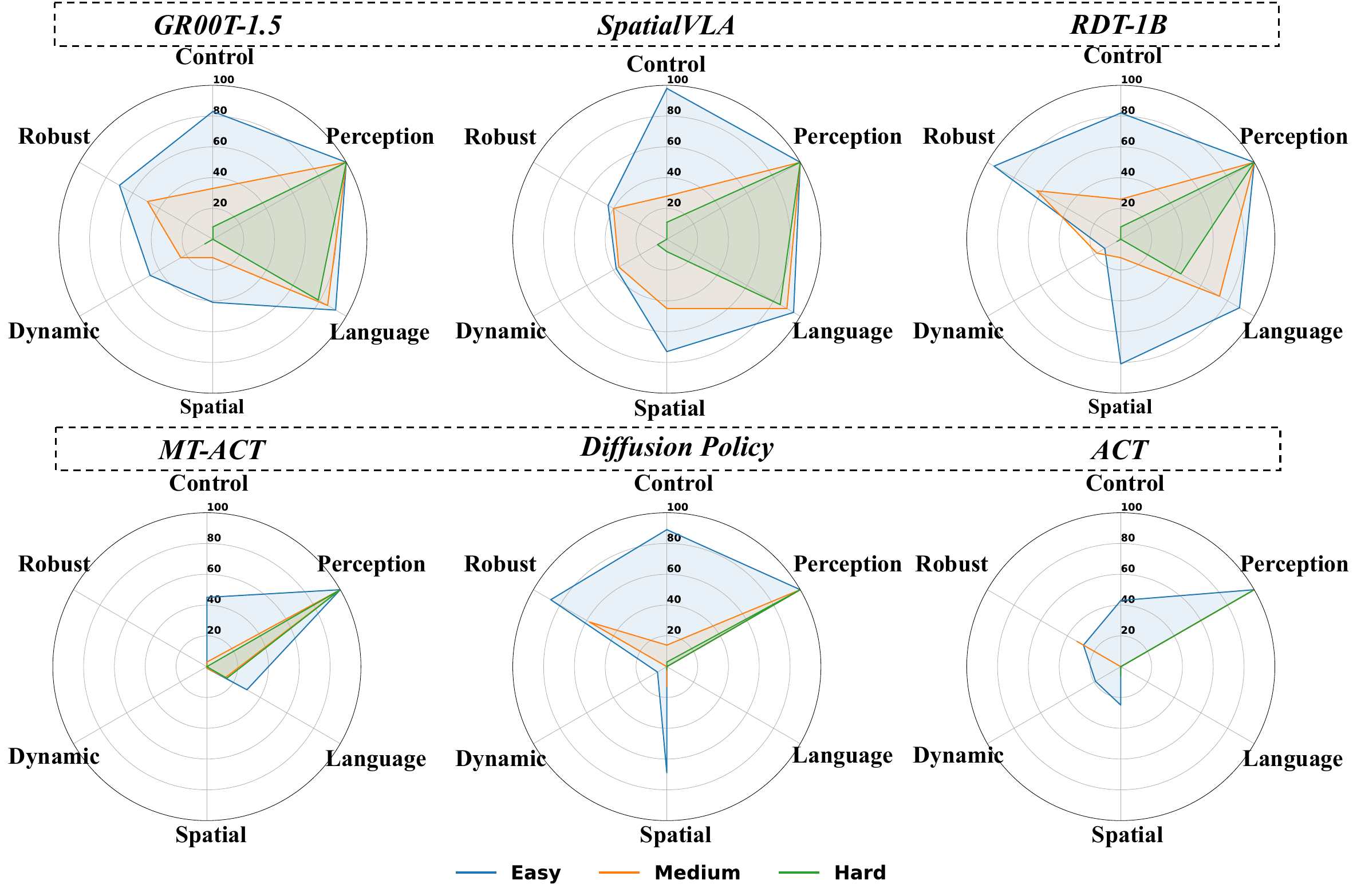}
    \caption{\textbf{Capability Radar Chart.} This figure presents a radar plot comparing the performance of evaluated policies across six core capability families in NEBULA: Perception, Control, Language, Spatial Reasoning, Dynamic Adaptation, and Robustness. Each axis shows the averaged success rate across task variants in each difficulty level. Higher values toward the outer edge indicate stronger performance in the isolated skill. The visualization reveals distinct strength–weakness profiles across models, highlighting complementary capabilities and critical failure modes.}
    \label{fig:capability_results}
\end{figure}

\vspace{-0.5em}
\section{Experiments}
To demonstrate the utility, coverage, and diagnostic strength of the NEBULA benchmark, we conduct comprehensive experiments across both capability and stress axes. These evaluations aim to answer several core questions: \emph{Can current embodied agents handle a wide range of skills?} \emph{Where do they fail under specific challenges?} And \emph{how can structured benchmarks help improve their design?} All experiments are conducted using the Alpha dataset to ensure consistency and reproducibility across tasks and conditions. This section focuses on the evaluated models (Section~\ref{subsec:baselines}) and their performance under the dual-axis evaluation framework (Section~\ref{sec:capabilityResults} and Section~\ref{sec:stressResults}).


\subsection{Baselines}
\label{subsec:baselines}
We evaluate a diverse set of state-of-the-art embodied agents to benchmark performance across NEBULA’s evaluation framework. Specifically, we include GR00T-1.5~\citep{bjorck2025gr00t}, SpatialVLA~\citep{qu2025spatialvla}, RDT-1B~\citep{liu2024rdt}, MT-ACT~\citep{bharadhwaj2024roboagent}, Diffusion Policy~\citep{chi2023diffusion}, and ACT~\citep{zhao2023learning}, which together represent a wide spectrum of architectural designs and control paradigms. For fair comparison, we unify data loading to match NEBULA’s format, keeping each model’s architecture, loss, and hyperparameters unchanged. All models are fine-tuned on NEBULA Alpha using their original training protocols.



\begin{figure}[!t]
    \centering
    \includegraphics[width=\textwidth]{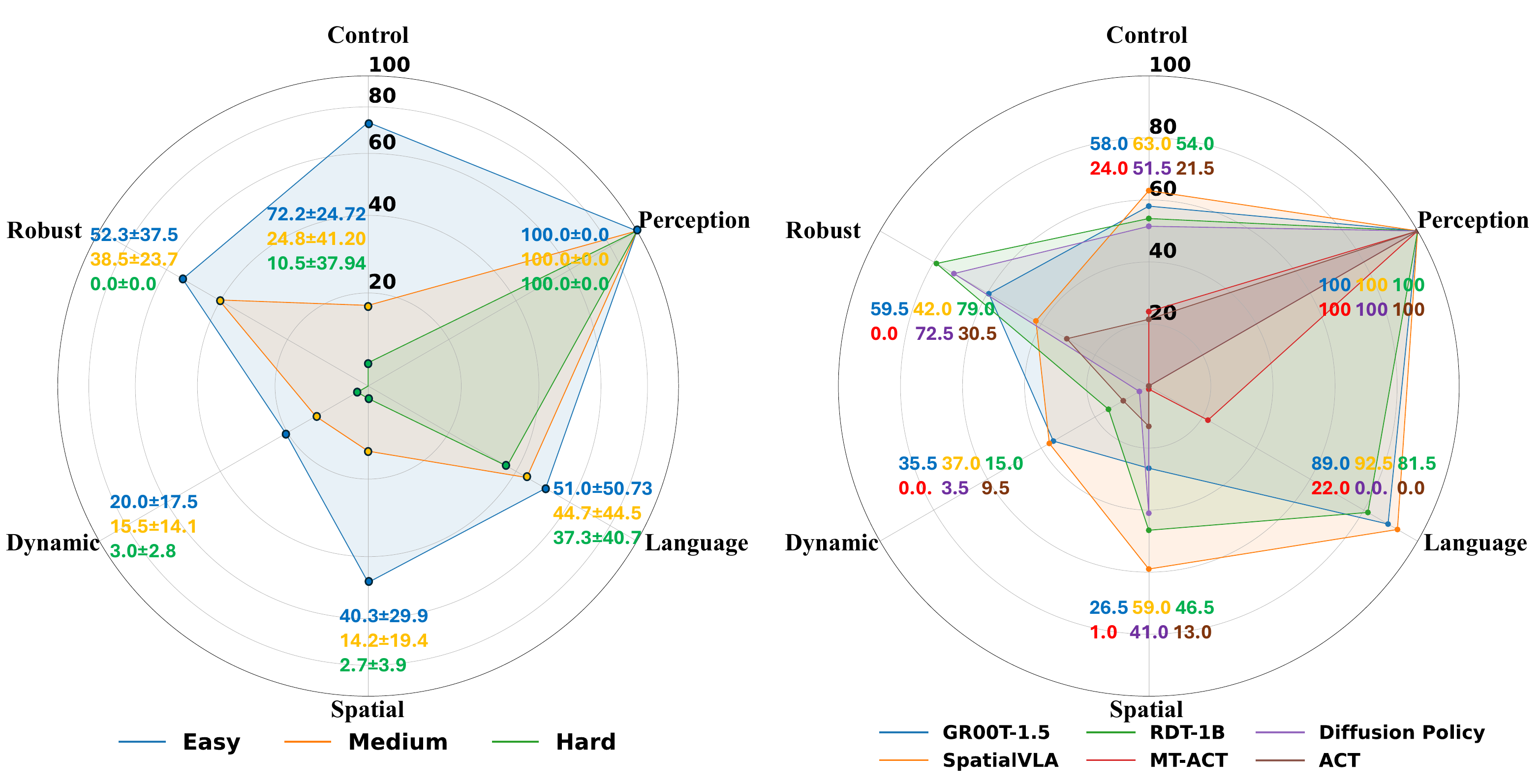}
    \caption{This figure presents two radar charts summarizing model performance across six capability task families. The \textbf{left} chart shows the mean $\pm$ standard deviation of success rates across all models for each task family at three difficulty levels. The \textbf{right} chart displays the average performance of individual models on Easy and Medium tasks, allowing for comparison across architectures. }
    \label{fig:overview_capability}
\end{figure}


\subsection{Capability Test Results}
\label{sec:capabilityResults}

As shown in Figure~\ref{fig:capability_results} and the left chart from Figure~\ref{fig:overview_capability}, the radar chart reveals several key trends in agent capabilities. Most models demonstrate strong performance in \textit{Perception} and \textit{Language} tasks. Nearly all baselines reliably identify object attributes like color and shape, even with distractors, indicating robust visual recognition. Similarly, these agents exhibit solid instruction grounding, successfully executing complex, conditional, and multi-step commands.

Performance on \textit{Control} and \textit{Spatial} tasks is more varied. SpatialVLA and GR00T-1.5 lead in \textit{Control}, handling long-horizon action sequences with high success. However, models like MT-ACT and ACT lag behind, revealing a need for better motor planning modules. \textit{Spatial reasoning} remains a key bottleneck for most models, with only SpatialVLA and RDT-1B achieving moderate success. Notably, even these models show clear drops from easy to medium level, especially under occlusion and containment conditions, indicating significant room for improvement in geometric reasoning.

All models struggle on \textit{Dynamic Adaptation} and \textit{Robustness} tasks, as shown in Figure~\ref{fig:overview_capability}. None of the evaluated VLA systems reliably adapts to time-sensitive triggers, distractors, or goal shifts, with radar scores near zero across the board. Similarly, robustness to distribution shifts(\eg novel object appearances, unseen layouts) is consistently poor, especially at higher difficulty levels. These results expose a major gap in current VLA capabilities and highlight two urgent research directions: real-time adaptive planning and out-of-distribution generalization.

We exclude the Hard level from the right radar chart in Figure~\ref{fig:overview_capability} because most models exhibit near-zero performance at this difficulty tier, especially in tasks involving robustness and dynamic adaptation. By focusing on Easy and Medium tasks, the chart provides clearer insights into current model capabilities. We hope this visualization encourages future research to close the gap at the Hard level, ultimately enabling more capable and resilient VLA systems.


\begin{figure}[!t]
    \centering
    \includegraphics[width=\textwidth]{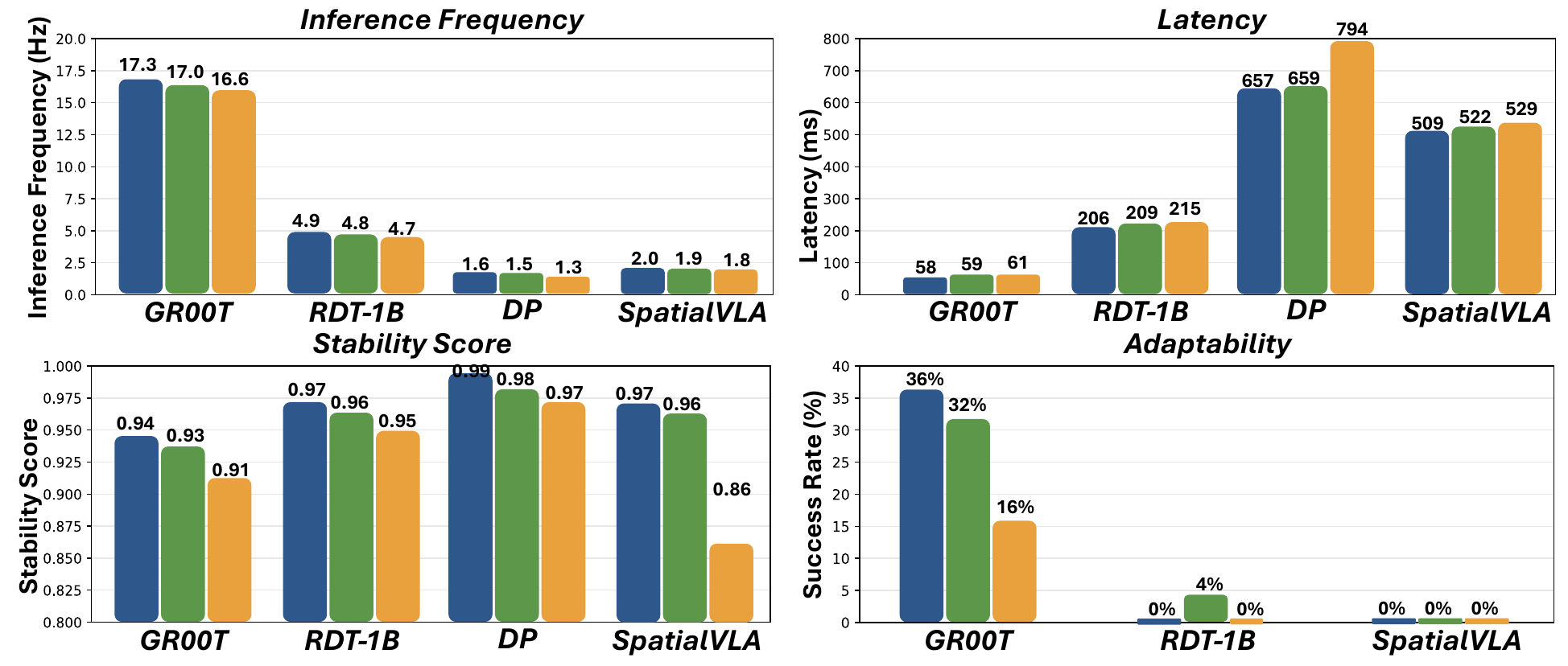}
    \caption{\textbf{Stress Test Evaluations.} This figure compares four models across three stress levels ($v1$, $v2$, $v3$), evaluating inference frequency (Hz), latency (ms), stability score (0–1), and adaptability. Higher values indicate better performance for all metrics except latency, where lower is preferred.}
    \label{fig:stress_results}
\end{figure}


\subsection{Stress Test Results}
\label{sec:stressResults}

As shown in Figure~\ref{fig:stress_results}, all evaluated models exhibit consistent performance degradation as stress levels increase, revealing sensitivity to deployment-time challenges such as computational bottlenecks and real-time demands. Inference frequency shows a clear decline across models: GR00T-1.5B remains the most resilient, maintaining ~$17Hz$ under $v3$, while DP and SpatialVLA fall below $2Hz$. This suggests many models struggle to meet real-time requirements under stress, and performance in ideal conditions may not generalize to practical deployment.

Latency results mirror this trend: with rising pressure, most models exhibit notable increases in response delay. Particularly, DP shows significant latency inflation, with its average step time nearly quadrupling from $v1$ to $v3$, peaking at around 800ms. This is indicative of inefficient model behavior under strain, potentially caused by unstable sampling mechanisms or computation-heavy policy architectures. In comparison, models like GR00T and RDT show slower rates of degradation, maintaining sub-300ms latency even under $v3$. These observations collectively highlight a key bottleneck for real-world deployment, where maintaining both throughput and response time is essential for safe and effective robot operation.

The stability score, measuring the smoothness of action trajectories (1.0 indicates perfect stability), also reveals growing fragility under stress. While RDT and DP maintain high scores above 0.95, SpatialVLA drops from 0.96 to 0.86 under $v3$, suggesting vulnerabilities in policy determinism. This decline may stem from increased decision-making stochasticity or sensitivity to input noise, leading to unreliable behaviors in dynamic scenarios where smooth, precise motion is essential.

Finally, the Adaptability results demonstrate that most current models are unable to handle dynamically evolving conditions. Except for GR00T, which shows modest success under adaptive task settings, all other models fail almost completely, with near-zero success rates across stress levels. This indicates a fundamental limitation in current VLA systems when faced with shifting goals, interactive feedback, or rapid environmental changes.

\begin{wraptable}{r}{0.5\linewidth}
\vspace{-5mm}
\caption{Success rates of GR00T-1.5 on three Perception (Easy level) tasks, comparing settings with isolated factors versus unisolated scenes with additional distractors. Results highlight the impact of scene confounding on perceptual accuracy.}
\vspace{-1em}
\centering
\Large
\setlength{\tabcolsep}{6pt} 
\renewcommand{\arraystretch}{1.0} 
\small
\resizebox{\linewidth}{!}{ 
\begin{tabular}{@{}cccc@{}}
\toprule
                                                           & \multicolumn{3}{c}{Perception (Easy Level)}                                                                                                                                         \\ \cmidrule(l){2-4} 
\begin{tabular}[c]{@{}c@{}}Isolated\\ Factors\end{tabular} & \begin{tabular}[c]{@{}c@{}}PlaceBigger \\ Sphere\end{tabular} & \begin{tabular}[c]{@{}c@{}}Place\\ Red Sphere\end{tabular} & \begin{tabular}[c]{@{}c@{}}Place\\ Sphere\end{tabular} \\ \midrule
\cmark                                                      & 100                                                           & 100                                                        & 100                                                    \\
\xmark                                                      & 92                                                            & 68                                                         & 76                                                     \\ \bottomrule
\end{tabular}
} 
\label{table:validity}
\vspace{-1.5em}
\end{wraptable}


\subsection{Validity of Factor Isolation}
We validate NEBULA’s factor isolation by comparing perception tasks in isolated vs. entangled settings. In isolation, the robot only needs to touch the correct object using simple language; in contrast, the entangled baseline requires full grasp-and-place execution involving multiple skills. As shown in Table~\ref{table:validity}, GR00T-1.5B achieves 100\% success in isolated settings, but drops to 92\%, 68\%, and 76\% when entangled. Video review shows failures are due to control and 3D spatial reasoning errors, highlighting how unrelated bottlenecks can obscure perception performance and validating NEBULA’s controlled-variable design.

%% file: sections/5-discussion.tex
\vspace{-0.5em}
\section{Discussion}

\begin{wraptable}{r}{0.4\linewidth}
\vspace{-4mm}
\caption{Success rates of different models on Robustness tasks, used to evaluate whether performance drop stems from high-level planning or low-level execution failures.}
\vspace{-1em}
\centering
\Large
\setlength{\tabcolsep}{6pt} 
\renewcommand{\arraystretch}{1.0} 
\small
\resizebox{\linewidth}{!}{ 
\begin{tabular}{@{}lcc@{}}
\toprule
\multicolumn{1}{c}{\multirow{2}{*}{Model}} & \multicolumn{2}{c}{Robust/Generalization}                                                                             \\ \cmidrule(l){2-3} 
\multicolumn{1}{c}{}                       & \begin{tabular}[c]{@{}c@{}}Easy\\ StackCube\end{tabular} & \begin{tabular}[c]{@{}c@{}}Medium\\ StackCube\end{tabular} \\ \midrule
PaliGemma                                  & 85                                                       & 75                                                         \\
SpatialVLA                                 & 0                                                        & 0                                                          \\ \midrule
Qwen                                       & 100                                                      & 90                                                         \\
GR00T-1.5                                  & 75                                                       & 0                                                          \\ \bottomrule
\end{tabular}
} 
\label{table:robust}
\vspace{-1.5em}
\end{wraptable}

\subsection{Why Are Generalization and Dynamic Performance Poor?}

To investigate why embodied agents struggle with generalization and dynamics, we decoupled the vision-language backbone from action head. Using SpatialVLA and GR00T's VLMs, we prompted high-level plans from static NEBULA scenes and had human annotators assess their validity.

As shown in Table~\ref{table:robust}, the standalone VLMs produce consistently valid strategies even in robustness tasks. However, their integrated VLA counterparts fail to execute these plans, with success rates dropping to zero under even moderate difficulty. This highlights a critical bottleneck: strong reasoning from VLMs does not guarantee successful embodied behavior, due to limitations in the action heads’ ability to translate abstract plans into precise control actions.

This issue is compounded by the inadequacy of conventional benchmarks that rely solely on success rate, obscuring whether failures arise from perception, reasoning, or control. NEBULA’s dual-axis evaluation addresses this by disentangling high-level reasoning from low-level execution and surfacing weaknesses under stress, offering the diagnostic granularity needed to build more robust and generalizable embodied systems.

\begin{wraptable}{r}{0.5\linewidth}
\vspace{-4mm}
\caption{Comparison of inference speed, latency, and adaptation score across models. Only GR00T-1.5 demonstrates meaningful adaptation, likely due to its significantly lower latency and higher inference frequency.}
\vspace{-1em}
\centering
\Large
\setlength{\tabcolsep}{6pt} 
\renewcommand{\arraystretch}{1.0} 
\small
\resizebox{\linewidth}{!}{ 
\begin{tabular}{@{}cccc@{}}
\toprule
Model      & \begin{tabular}[c]{@{}c@{}}Avg.\\ Inference \\ Frequency\end{tabular} & \begin{tabular}[c]{@{}c@{}}Avg.\\ Latency\end{tabular} & \begin{tabular}[c]{@{}c@{}}Avg.\\ Adaptation\end{tabular} \\ \midrule
GR00T-1.5  & 16.98 Hz                                                              & 58.62 ms                                               & 28                                                        \\
RDT-1B     & 4.84 Hz                                                               & 206.77 ms                                              & 1                                                         \\
SpatialVLA & 1.92 Hz                                                               & 520.48 ms                                              & 0                                                         \\ \bottomrule
\end{tabular}
} 
\label{table:responseVSdynamic}
\vspace{-1.5em}
\end{wraptable}

\vspace{-1em}
\subsection{Fast Inference Is Key to Dynamic Adaptation}
As shown in the Capability Test (Figure~\ref{fig:overview_capability}), nearly all models fail to handle Dynamic tasks. To further investigate this weakness, we introduced an Adaptation Stress Test to simulate dynamic environments and evaluate model robustness under goal shifts and real-time disruptions. Table~\ref{table:responseVSdynamic} compares inference frequency, latency, and adaptation success across models to help uncover underlying causes.

Results show that GR00T-1.5 is the only model with moderate adaptation (success rate = 28), while RDT-1B and SpatialVLA perform poorly. GR00T-1.5 also demonstrates the fastest response—16.98 Hz inference frequency and 58.62 ms latency—whereas slower models show little adaptive behavior. This suggests that fast perception and replanning are crucial for dynamic environments.

Overall, these findings highlight a critical bottleneck: real-time adaptation requires not only high-level reasoning but also fast, low-latency control pipelines. Future work should focus on optimizing system responsiveness—especially in the action head—to enable effective online replanning and robust behavior under dynamic conditions.




%% file: sections/6-conclusion.tex
\vspace{-0.5em}
\section{Conclusion}
In this work, we introduced NEBULA, an evaluation-first ecosystem that unifies fragmented data formats and establishes a dual-axis framework for embodied AI. By disentangling capability tests from stress tests and enforcing controlled-variable isolation, NEBULA provides interpretable, skill-specific, and robust performance diagnostics that go beyond conventional success rates. Our experiments demonstrate how this design reveals hidden bottlenecks, clarifies model strengths and weaknesses, and lays the foundation for systematic progress toward reliable VLA agents.

%% file: sections/appendix_tasks.tex
\subsection{Unified Data Platform}
\label{appendix:unified_data_platform}
This section details the NEBULA Unified Data Platform, a comprehensive ecosystem designed to address the severe data fragmentation that hinders progress in embodied AI research. The current landscape, characterized by numerous benchmarks with disparate and incompatible data formats, forces researchers to reimplement data processing pipelines for each new dataset. This fundamental challenge not only prevents fair head-to-head model comparisons but also limits the large-scale generalization studies necessary for advancing the field. Our platform overcomes this bottleneck by unifying these varied data sources into a modular and extensible interface. It provides a structured, robot-agnostic foundation for working with large-scale VLA datasets, thereby establishing the necessary infrastructure for the fair and scalable evaluation our work introduces.

The platform’s architecture is founded on two core design principles that ensure both consistency and extensibility. The first is its Unified and Structured Episode Format, which establishes a canonical, robot-agnostic data structure for representing any robot interaction trajectory. An Episode is defined as one complete task attempt, containing a language instruction, a time-ordered sequence of Steps, and comprehensive metadata. Each Step encapsulates the system's state at a discrete timestep, comprising a multi-modal Observation (including multiple camera views and proprioceptive states) and the corresponding Action taken by the agent. The second principle is a Robot-Abstracted Embodiment Layer, which decouples robot-specific properties from the core data logic. Instead of being hardcoded, hardware characteristics such as degrees of freedom, gripper types, and multi-arm configurations are defined in a centralized configuration system, making the platform inherently extensible to new robotic hardware with minimal effort.

This architecture is made accessible to researchers through a high-level Python Software Development Kit (SDK) designed to streamline the research workflow. The SDK abstracts away the complexities of file discovery, parsing, and data decoding, providing a clean interface for programmatic access. Its central feature is a powerful, fluent query engine that allows researchers to efficiently filter and sample data based on a wide range of metadata attributes, including task family, success status, trajectory length, or even natural language instructions. To further support robust and reproducible experimentation, the platform also includes built-in utilities for common machine learning workflows, such as stratified train-test splitting, which ensures that model validation is performed on balanced and representative data subsets. 

Ultimately, the NEBULA Unified Data Platform makes a critical contribution to the field by removing the significant engineering overhead associated with data fragmentation. This allows the research community to shift its focus from the tedious work of data wrangling to the core challenges of model innovation and architectural design. More importantly, this unified infrastructure is the essential foundation upon which our dual-axis evaluation framework is built. By ensuring data consistency and enabling plug-and-play compatibility with a wide range of models , the platform provides the necessary conditions for the fair, large-scale, and diagnostic evaluations required to systematically advance the development of robust and generalizable embodied agents.

\subsection{Test Tasks Design \& Evaluation}
\label{appendix:tasks_details}


\subsubsection{Capability Test Tasks}
\label{appendix:capabilty}
This section provides the full specification for all capability tasks used in NEBULA. Each task family targets a specific embodied competency and includes \textit{Easy}, \textit{Medium}, and \textit{Hard} tiers. Each task tier comprises three unique task sets instantiated from templates, with randomized object attributes and layout to ensure diversity. Success is determined through well-defined, programmatic criteria based on object positions, interactions, and task logic.

\begin{figure}[!t]
    \centering
    \includegraphics[width=\textwidth]{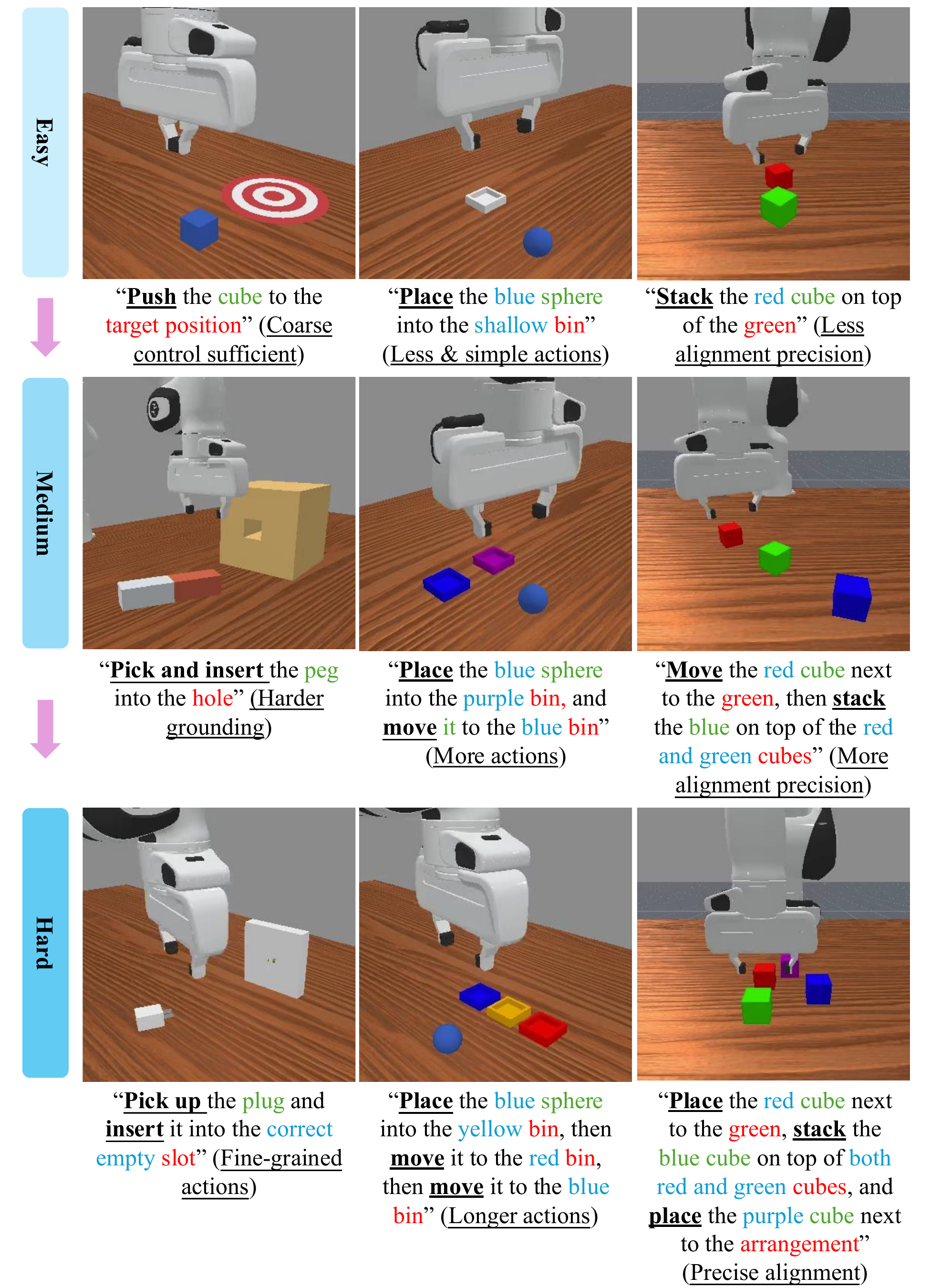}
    \vspace{-2em}
    \caption{The \textbf{Control capability} family evaluates an agent’s ability to perform precise and reliable motor actions under varying levels of complexity. \textcolor{green}{Green} marks objects, \textcolor{red}{red} marks targets, and \textcolor{blue}{blue} indicates contextual cues. \textbf{\underline{Bold underlined}} text shows actions; \textit{\underline{italic underlined}} text gives clarifications.}
    \label{fig:control_tasks}
    \vspace{-2em}
\end{figure}

\paragraph{Control} The Control task family is specifically designed to isolate an agent's low-level manipulation and motion planning capabilities by systematically removing all non-control-related confounding factors. To ensure this isolation, each task instance provides fully specified and fixed object states, including fixed positions, orientations, and visual properties such as color and size. This eliminates any reliance on perception, semantic understanding, or language grounding. Instructions are deliberately minimal and unambiguous, ensuring that task success depends purely on motor execution. 

The task suite is divided into three difficulty tiers based on control sequence complexity. \textit{Easy} tasks involve one to two atomic actions such as picking and pushing. Success is determined based on the object reaching its goal position under proper spatial relation. \textit{Medium} tasks extend to multiple sequential steps, often requiring coordination across multiple objects, with success requiring completion of the full action sequence with all spatial constraints satisfied. \textit{Hard} tasks require extended action sequences involving more steps and include fine-grained manipulation challenges that demand high precision and stability, such as multi-object stacking arrangements or sub-centimeter insertion tolerances. This progression allows for a nuanced evaluation of an agent's capacity to generate, maintain, and adjust action sequences in increasingly demanding scenarios. The visualization and corresponding language commands are demonstrated in Figure \ref{fig:control_tasks}.

\begin{figure}[!t]
    \centering
    \includegraphics[width=\textwidth]{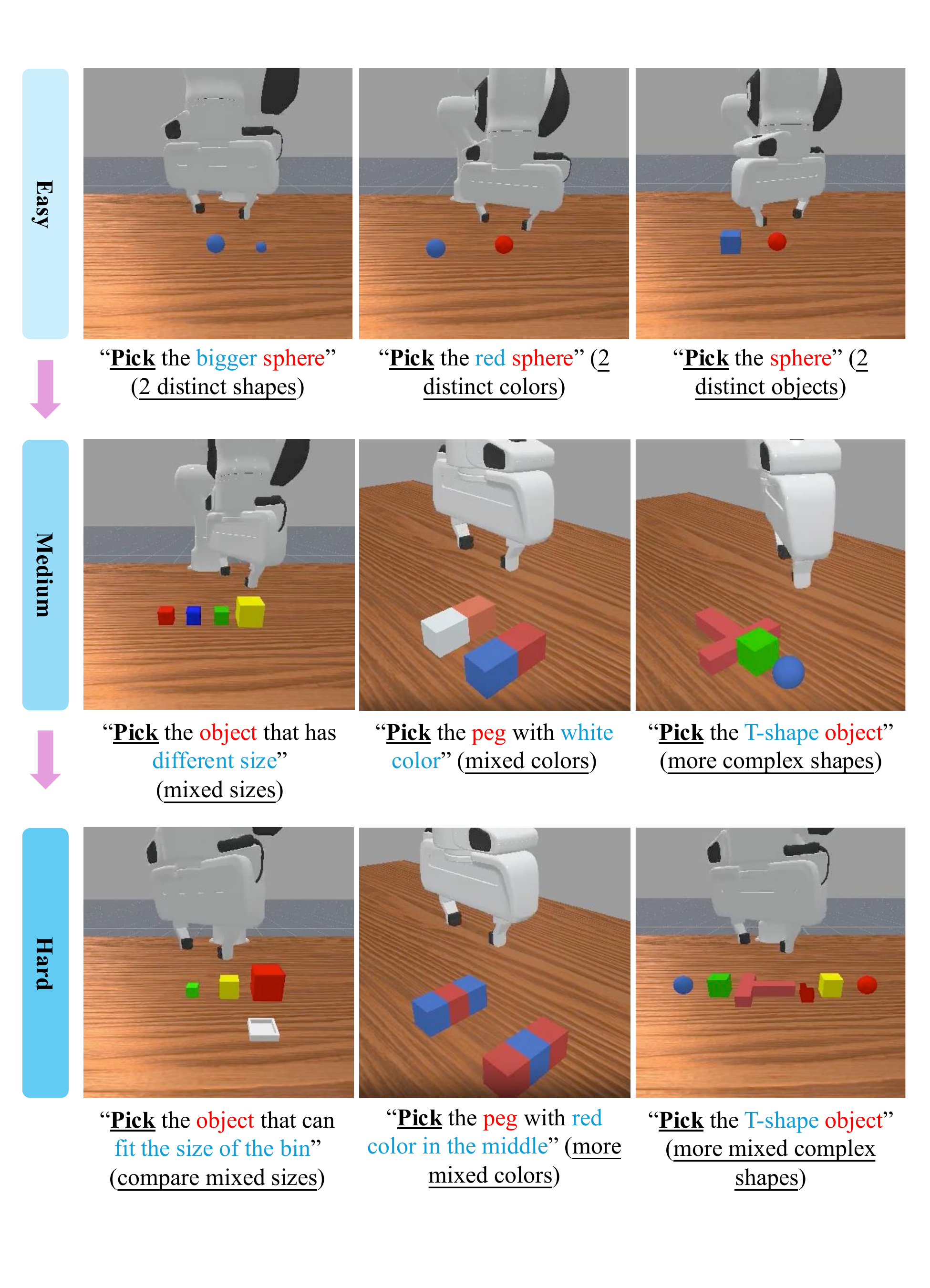}
    \vspace{-5em}
    \caption{The \textbf{Perception capability} test is designed to evaluate an agent’s ability to identify target objects based solely on their visual attributes. \textcolor{red}{red} marks targets, and \textcolor{blue}{blue} indicates contextual cues. \textbf{\underline{Bold underlined}} text shows actions.}
    \label{fig:perception_tasks}
    \vspace{-2em}
\end{figure}

\paragraph{Perception} The Perception task family evaluates an agent’s ability to recognize and distinguish object-level attributes such as color, shape, and size, while explicitly eliminating confounding factors from downstream control. To ensure a clean probe into perceptual capacity, control difficulty is minimized: all target objects are placed within reachable, uncluttered regions, and the task is considered successful as long as the robot makes contact with the correct object—regardless of grasp success or trajectory smoothness. This design mitigates potential confounds from hardware instability and isolates perception as the sole bottleneck.

Task difficulty increases across three tiers: \textit{Easy} tasks involve clearly distinct attributes; \textit{Medium} tasks introduce ambiguity via subtle shape or color variations across multiple distractors; and \textit{Hard} tasks incorporate partial occlusions and low-contrast distractors, requiring the agent to resolve fine-grained visual distinctions under constrained viewpoints. This progression enables robust evaluation of perceptual skills under increasingly realistic and complex visual conditions. The visualization and corresponding language commands are demonstrated in Figure \ref{fig:perception_tasks}.

\begin{figure}[!t]
    \centering
    \includegraphics[width=\textwidth]{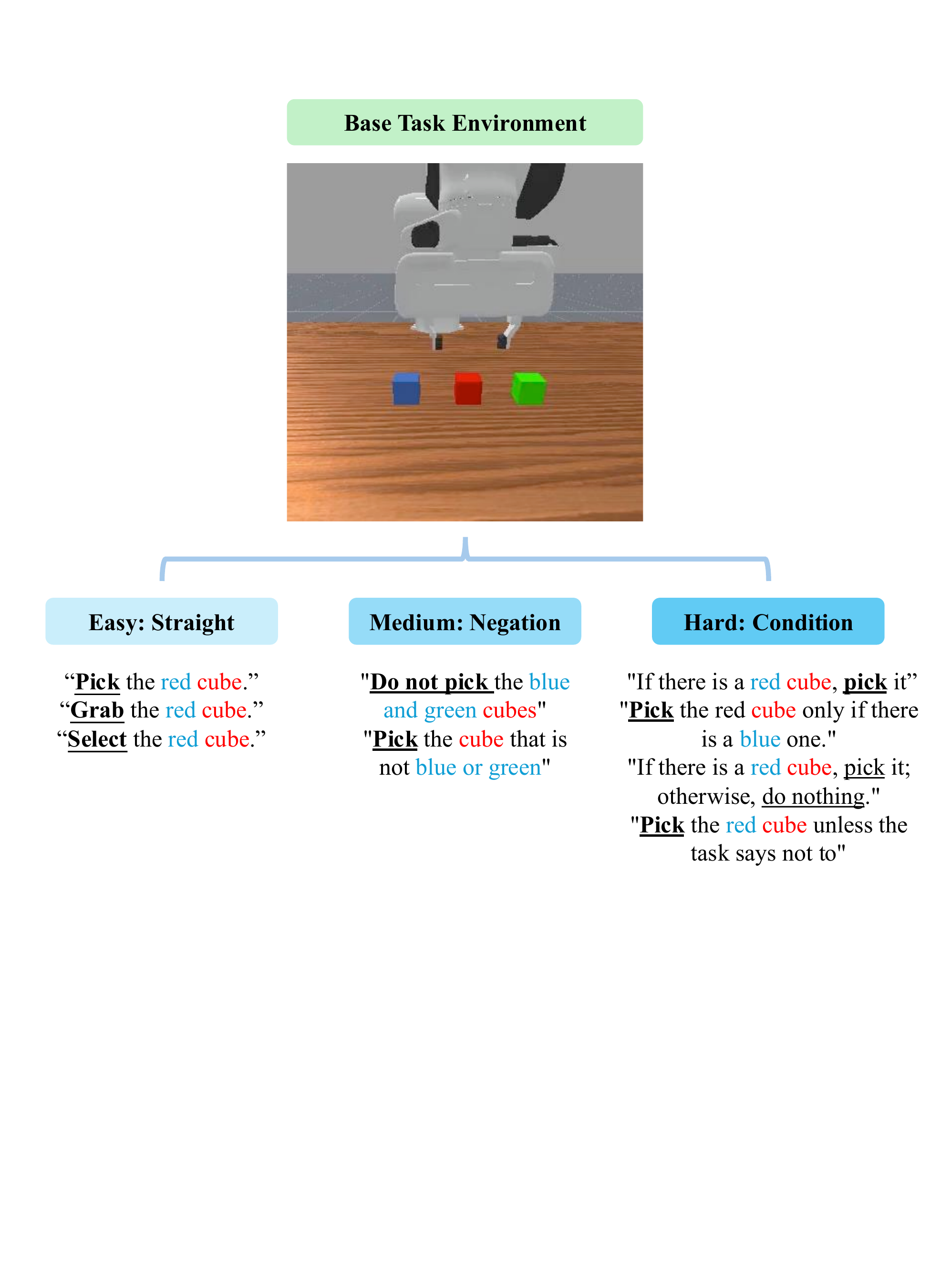}
    \vspace{-19em}
    \caption{The \textbf{Language capability} in NEBULA evaluates a model’s ability to interpret and act upon natural language instructions in robotic manipulation settings. \textcolor{red}{red} marks targets, and \textcolor{blue}{blue} indicates contextual cues. \textbf{\underline{Bold underlined}} text shows actions.}
    \label{fig:language_tasks}
    \vspace{-1em}
\end{figure}

\paragraph{Language} The Language task family is designed to isolate and evaluate an agent’s ability to interpret natural language instructions with minimal interference from perception, control, or environmental variability. To enforce this isolation, all scenes are fully standardized across difficulty levels—identical objects, visual attributes, and spatial configurations are used for every task variant. Only the instruction text changes, ensuring that observed performance differences stem solely from linguistic understanding rather than scene-specific cues or motor complexity. 

The tasks are categorized into three difficulty tiers: \textit{Easy} tasks test basic grounding of surface-level attributes (\eg “Pick the red cube”); \textit{Medium} tasks require relative position analysis and selective instruction comprehension (\eg “Place the cube that is not red” or “Pick the small green cube”); and \textit{Hard} tasks assess deeper linguistic reasoning, including conditional logic, instruction filtering, and multi-step execution tracking (\eg “If the green cube is smaller than the red one, place it in the bin. Otherwise, discard it”). This setup provides a clean and controlled probe into the agent’s ability to parse, interpret, and act upon language-based directives of increasing semantic and logical complexity. The visualization and corresponding language commands are demonstrated in Figure \ref{fig:language_tasks}.

\begin{figure}[!t]
    \centering
    \includegraphics[width=\textwidth]{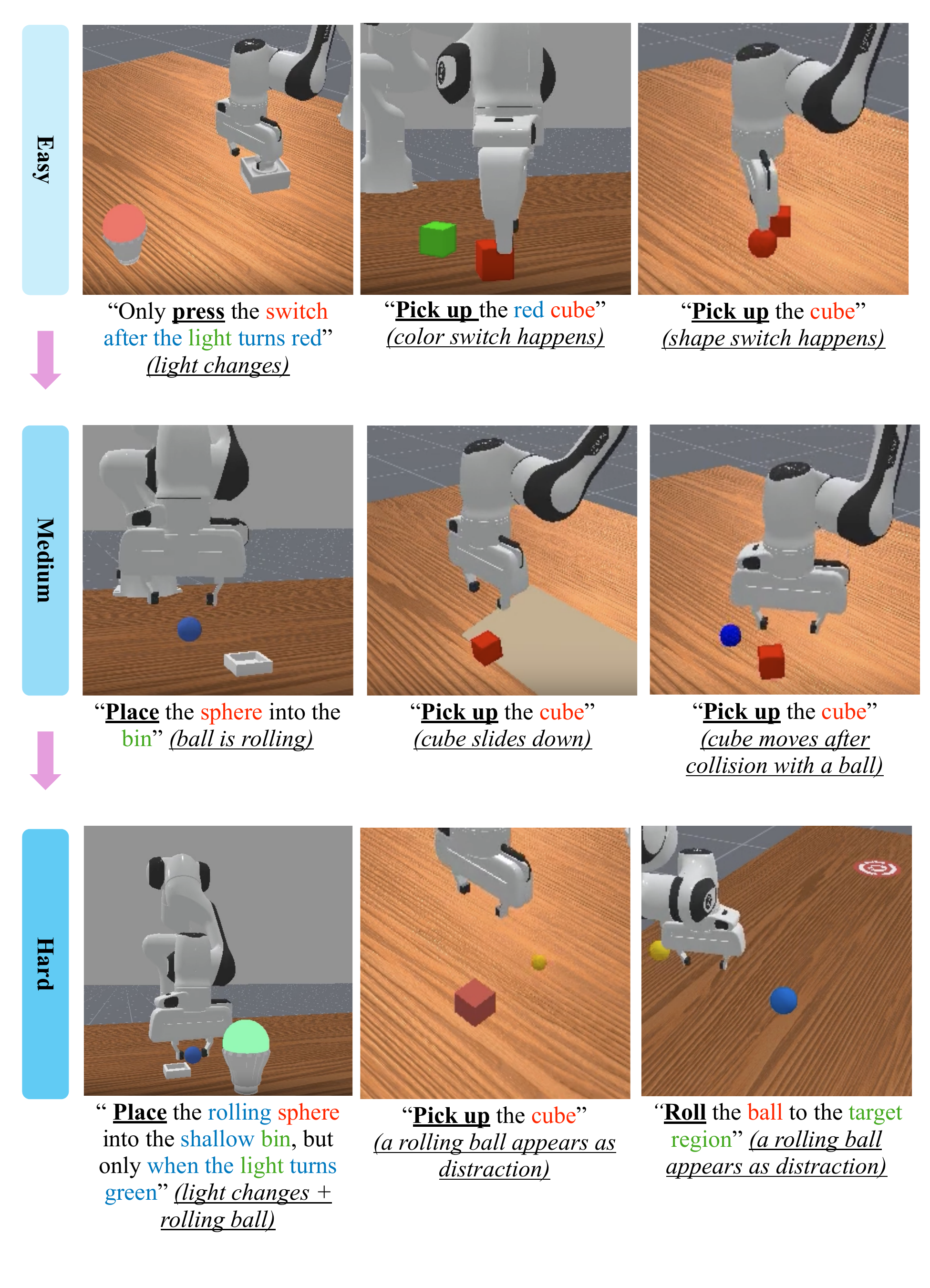}
    \vspace{-3.5em}
    \caption{The \textbf{Dynamic Adaptation capability} tests in NEBULA are designed to evaluate an agent’s ability to perceive and respond to changes in the environment in real time. \textcolor{green}{Green} marks objects, \textcolor{red}{red} marks targets, and \textcolor{blue}{blue} indicates contextual cues. \textbf{\underline{Bold underlined}} text shows actions; \textit{\underline{italic underlined}} text gives clarifications. }
    \label{fig:dynamic_tasks}
\end{figure}

\paragraph{Dynamic Adaptation} The Dynamic Adaptation task family targets an agent’s ability to operate under time-varying and non-stationary conditions, evaluating how well it can adjust to moving objects, time-sensitive constraints, and external perturbations. 

This task family is structured into three difficulty tiers, each progressively increasing the level of environmental dynamics and required reactivity. \textit{Easy} tasks involve static scenes with time-critical or distraction-based events, such as pressing a switch within a short time window. \textit{Medium} tasks introduce slow, predictable dynamics in the scene—objects may roll, slide, or shift position over time, requiring the agent to adjust its plan on-the-fly. These tasks require basic perception-action adaptation and temporal anticipation. \textit{Hard} tasks present high-variability and multi-modal dynamics. These tasks demand complex real-time perception, state tracking, and policy re-evaluation, pushing the limits of an agent’s reactive robustness and memory. By scaling the difficulty along temporal variability and unpredictability, this task family offers a comprehensive stressor for evaluating embodied agents in non-static environments. The visualization and corresponding language commands are demonstrated in Figure \ref{fig:dynamic_tasks}.

\begin{figure}[!t]
    \centering
    \includegraphics[width=\textwidth]{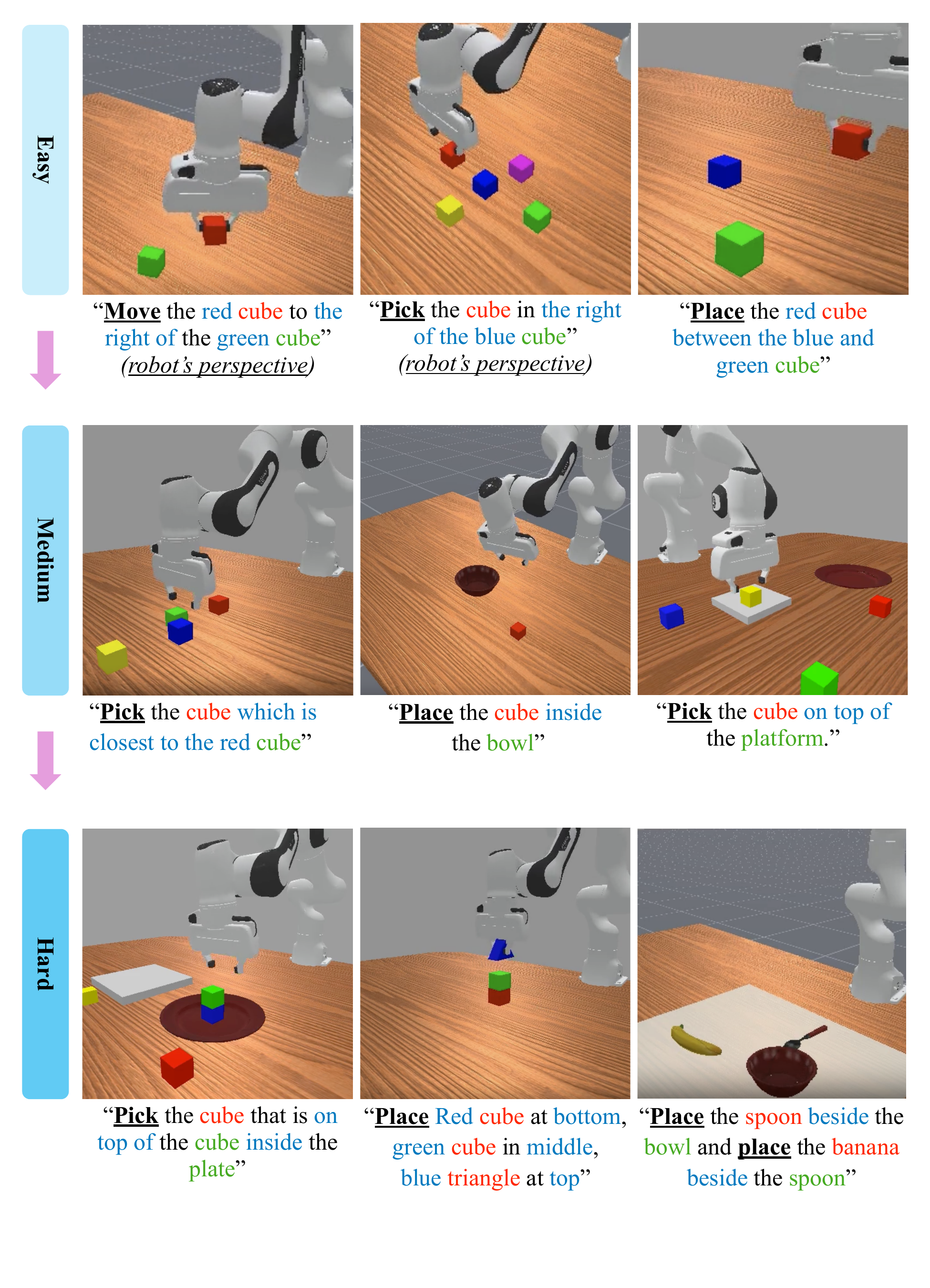}
    \vspace{-5em}
    \caption{The \textbf{Spatial Reasoning capability} family evaluates an agent's ability to interpret and execute spatial relationships and geometric constraints in 3D manipulation tasks. \textcolor{green}{Green} marks objects, \textcolor{red}{red} marks targets, and \textcolor{blue}{blue} indicates contextual cues. \textbf{\underline{Bold underlined}} text shows actions; \textit{\underline{italic underlined}} text gives clarifications. }
    \label{fig:spatial_tasks}
    \vspace{-1em}
\end{figure}

\paragraph{Spatial Reasoning} The Spatial task family evaluates an agent’s ability to reason over object positions and geometric relationships in 3D space. Unlike perception tasks that focus on attribute recognition, these tasks isolate spatial understanding by holding visual appearance and control difficulty fixed, ensuring that success depends solely on interpreting and executing spatial constraints. 

\textit{Easy} tasks are confined to 2D planar reasoning where relations like left, right, or between are defined on a flat surface. \textit{Medium} tasks expand to 3D spatial concepts, introducing both horizontal and vertical relationships.” \textit{Hard} tasks demand full 6-DoF motion planning, requiring the agent to align and manipulate objects in all three spatial axes with rotational precision, such as stacking irregularly shaped items in complex orientations. This progression enables controlled, fine-grained evaluation of spatial reasoning skills across increasing geometric complexity. The visualization and corresponding language commands are demonstrated in Figure \ref{fig:spatial_tasks}.

\begin{figure}[!t]
    \centering
    \includegraphics[width=\textwidth]{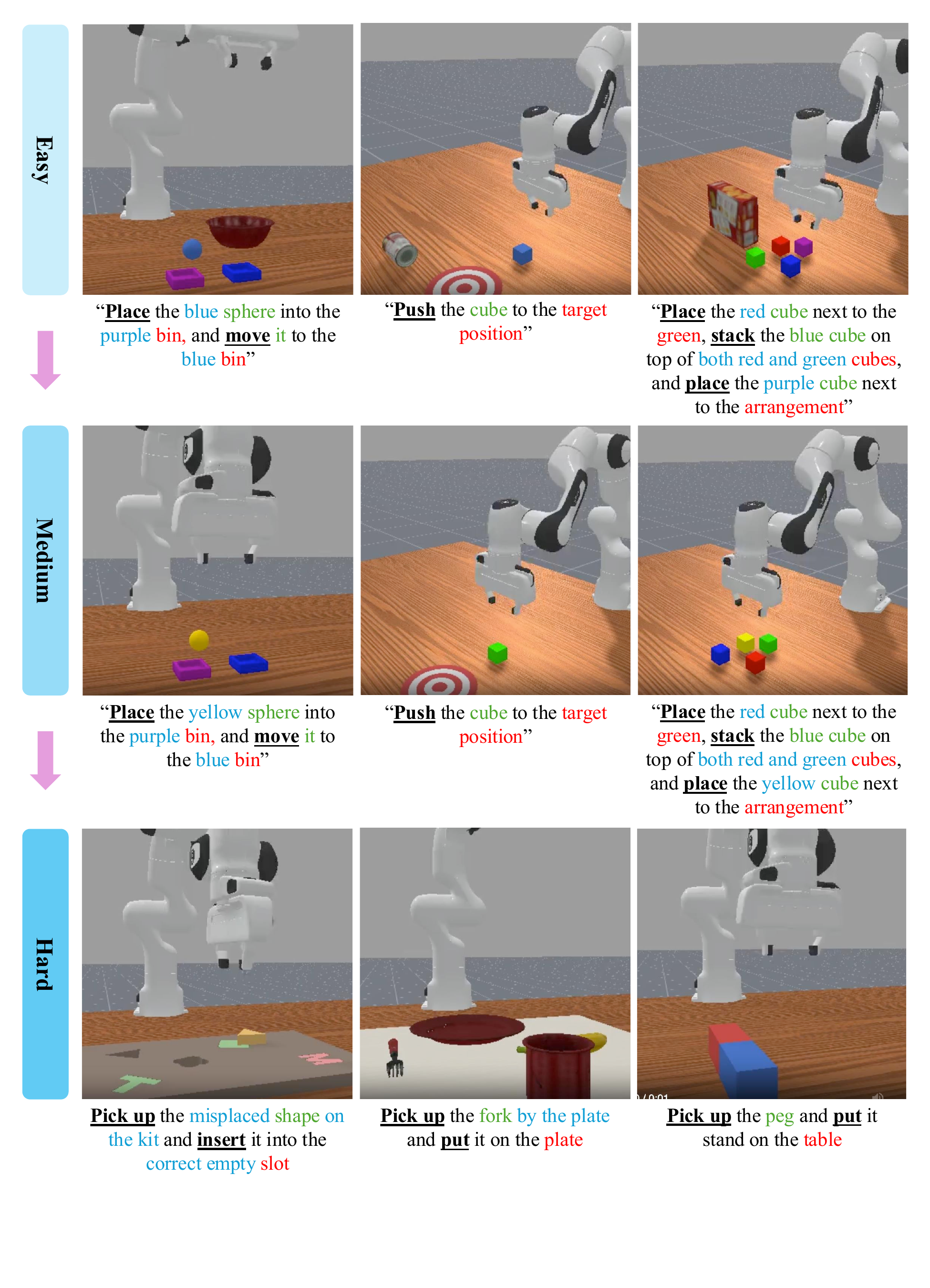}
    \vspace{-6em}
    \caption{The \textbf{Robustness/Generalization} capability in NEBULA evaluates an agent’s ability to perform reliably across diverse, unseen conditions. Tasks in this category are intentionally designed to expose the agent to variations it has not encountered during training.  \textcolor{green}{Green} marks objects, \textcolor{red}{red} marks targets, and \textcolor{blue}{blue} indicates contextual cues. \textbf{\underline{Bold underlined}} text shows actions.}
    \label{fig:robust_tasks}
    \vspace{-1.5em}
\end{figure}

\paragraph{Robustness/Generalization} The Robustness and Generalization task family is designed to assess an agent's ability to perform reliably under distribution shifts in object attributes, scene composition, and out-of-distribution (OOD) scenarios. 

\textit{Easy} tasks introduce distractor objects (1-2 objects) into familiar scenes while keeping the main task unchanged, testing an agent's selective attention. \textit{Medium} tasks alter object attributes such as color—for instance, changing the sphere from blue to orange while maintaining the same scene structure and task requirements—probing the agent's ability to generalize across visual variations. \textit{Hard} tasks present completely novel environments, layouts, or object configurations that were never encountered during training, thereby measuring the agent's generalization capacity to OOD scenarios. The success criteria are the same as the original task or are evaluated according to the correct spatial relations, positions, and orientations. Together, these progressively challenging setups evaluate how well embodied agents can adapt their learned policies to unfamiliar or perturbed conditions without retraining or explicit guidance. The visualization and corresponding language commands are demonstrated in Figure \ref{fig:robust_tasks}.


\subsubsection{Stress Test Tasks}
\label{appendix:stress}

\begin{figure}[!t]
    \centering
    \includegraphics[width=\textwidth]{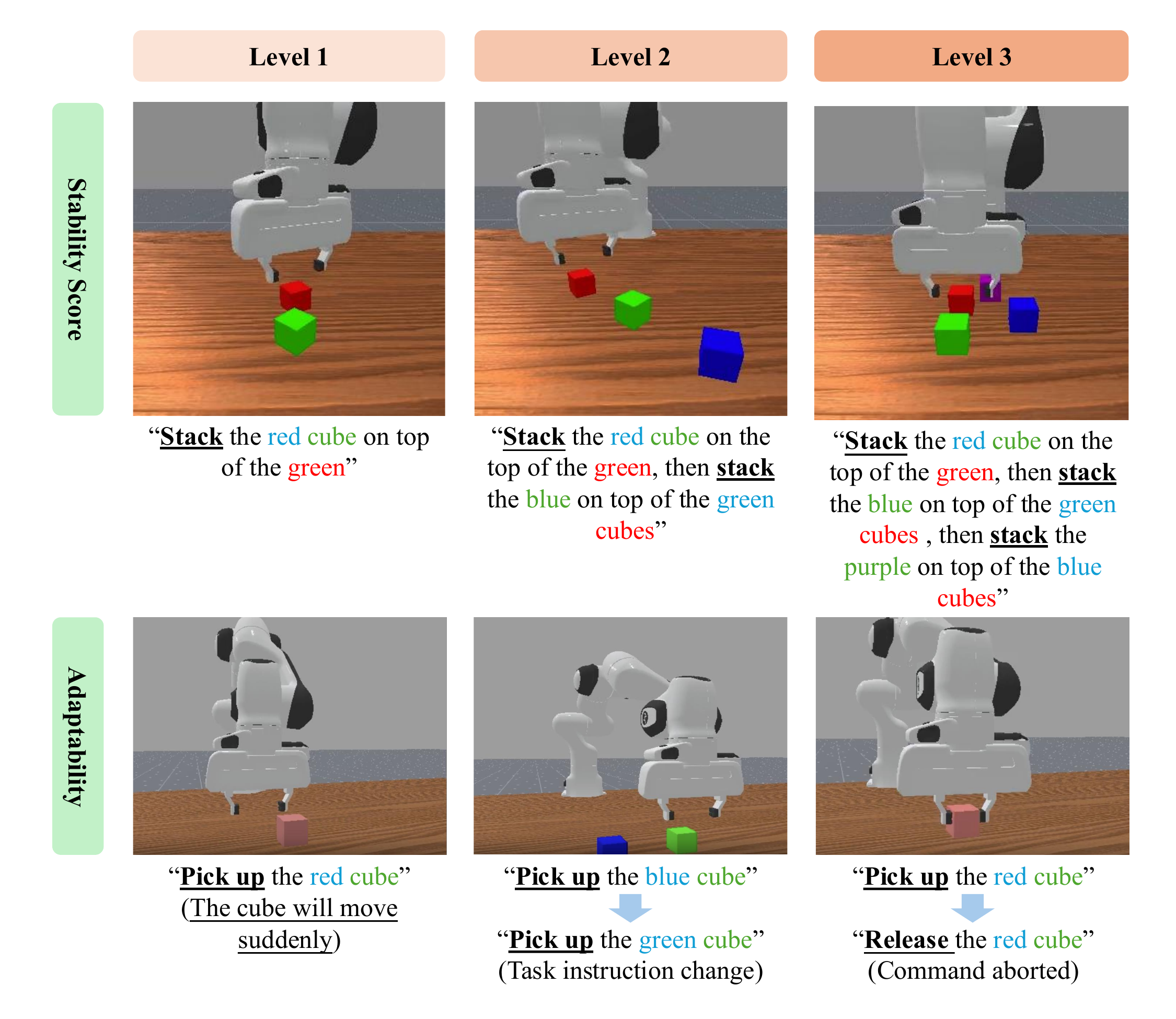}
    \vspace{-2em}
    \caption{Visualization of NEBULA \textbf{Stress Test}. \textcolor{green}{Green} marks objects, \textcolor{red}{red} marks targets, and \textcolor{blue}{blue} indicates contextual cues. \textbf{\underline{Bold underlined}} text shows actions; \textit{\underline{italic underlined}} text gives clarifications.}
    \label{fig:adapt_tasks}
    \vspace{-1em}
\end{figure}

This section presents the stress test specifications for evaluating system performance under operational constraints. Each test systematically varies a single performance indicator across three calibrated levels, with specific criteria detailed below.

\paragraph{Inference Frequency} Inference frequency measures the rate at which an agent generates control actions in hertz. This metric directly impacts an agent's ability to respond to dynamic environments and maintain smooth control. The test evaluates inference frequency under three scenarios: $v_1$ tests slow and uniform movements; $v_2$ tests alternating movement at medium speed; $v_3$ tests fast irregular movements. Performance degradation across tiers reveals how VLA models handle increasing computational demands, exposing whether failures stem from insufficient inference speed and model architecture limitations.

\paragraph{Latency} Latency quantifies the delay between sensory input and action output, measured in milliseconds from perception trigger to control signal generation. esting occurs across three conditions: $v_1$ measures static scene; $v_2$ measures dynamic scene with moving objects; $v_3$ measures dynamic scene with fast-moving objects. This metric is critical for time-sensitive manipulation where delayed responses lead to task failure. Lower latency enables tighter control loops and more responsive behavior, particularly crucial for contact-rich manipulation and dynamic grasping tasks.

\paragraph{Stability Score} Stability scores quantifies trajectory smoothness by measuring action variation between consecutive timesteps. Given an action sequence $\{a_{0}, a_{1}, ..., a_{t}\}$, the score is computed as:

$$\text{Stability} = \exp\left(-\frac{1}{T-1}\sum_{t=1}^{T}||\mathbf{a}_t - \mathbf{a}_{t-1}||_2\right)$$

where $||\mathbf{a}_t - \mathbf{a}_{t-1}||_2$ represents the $L2$ norm of action changes between neighboring timesteps and the exponential decay of mean action changes yields a normalized score $\in [0,1]$, with 1 indicating perfect stability. The test evaluates three precision levels: $v_1$ tests coarse continuous force control such as object pushing; $v_2$ requires smoother and more accurate trajectories like grasping and lifting operations; $v_3$ demands high-precision position and orientation control for tasks like plug insertion. This metric reveals whether VLA policies generate stable control signals suitable for physical deployment, distinguishing smooth execution from erratic behaviors that could damage hardware or cause task failure. Figure~\ref{fig:adapt_tasks} shows the task design.

\paragraph{Adaptability} Adaptability measures an agent's ability to adjust its behavior in response to environmental changes, task interruptions, or modified objectives during execution. The test evaluates the model's performance across three scenarios: $v_1$ tests response to object displacement where the target suddenly moves to a new position; $v_2$ introduces mid-task instruction changes, requiring the agent to switch between objectives (\eg "Pick up the blue cube" $\to$ "Pick up the green cube"); $v_3$ demands rapid re-planning under sequential instructions (\eg "Pick up the cube" $\to$ "Release the cube"). This progression assesses whether VLA policies can maintain task coherence under dynamic conditions, distinguishing reactive agents that gracefully handle perturbations from rigid controllers that fail when initial assumptions are violated. The visualization and corresponding language commands are demonstrated in Figure \ref{fig:adapt_tasks}.

\paragraph{Resources} The resources stress test evaluates the computational efficiency and scalability of embodied agents by measuring runtime resource consumption across multiple dimensions as well as the static memory usage. This test quantifies GPU memory usage, CPU memory usage, and model size. By profiling memory footprint and model size alongside task performance, this test enables practitioners to assess deployment feasibility across hardware-constrained platforms and identify computational bottlenecks that may limit real-world applicability.

\subsection{Benchmark Comparison}


\begin{table*}[t!]
\centering
    \caption{Comparison of NEBULA and existing single-arm manipulation benchmarks across task design and evaluation protocols. NEBULA uniquely supports both capability evaluation and stress testing. Unlike prior benchmarks, it adopts a dual-axis protocol that evaluates skills and stress responses separately, ensuring each score reflects a specific factor. Other benchmarks mostly report task-level success rate without isolating capabilities or stress conditions, limiting diagnostic insight.}
\resizebox{\linewidth}{!}{
\begin{tabular}{lcccccc}
\hline
\multirow{2}{*}{Benchmark} & \multicolumn{4}{c}{Task Design}                                               & \multicolumn{2}{c}{Data Design} \\ \cline{2-7} 
                           & Task Families    & Language              & Tiered Difficulty     & Evaluation & \# Modality      & \# View      \\ \midrule
ManiSkill                  & Multiple         & \xmark & \xmark & TSR        & 1                & 3            \\
RLBench                    & Multiple         & \xmark & \xmark & TSR        & 1                & 2            \\
FurnitureBench             & Furniture        & \xmark & \xmark & TSR        & 1                & 2            \\
BridgeDataV2               & Pick/Place       & \xmark & \xmark & TSR        & 1                & 2            \\
Meta-World                 & Multiple         & \xmark & \xmark & TSR        & 1                & 2            \\
FrankaKitchen              & Kitchen-related  & \xmark & \xmark & TSR        & 1                & 2            \\
CLVIN                      & Visual Reasoning & \cmark & \xmark & TSR        & 1                & 2            \\
ALFRED                     & Compositional    & \cmark & \xmark & TSR        & 1                & 2            \\
LIBERO                     & Language         & \cmark & \cmark & TSR        & 1                & 2            \\
VLABench                   & Realistic        & \cmark & \cmark & TSR        & 1                & 2            \\ \midrule
\textbf{NEBULA (Ours)}     & 6 Capabilities   & \cmark & \cmark & DAE        & 3                & 6            \\ \bottomrule
\end{tabular}
    }
    \caption*{(\textbf{Notes}: \emph{SR} represents \emph{Success Rate}, \emph{TSR} represents \emph{Task-level Success Rate}, \emph{DAE} represents \emph{Dual-Axis Evaluation})}
    \label{table:nebulaVSbenchmark}
    \vspace{-6mm}
\end{table*}

Table \ref{table:nebulaVSbenchmark} indicates that NEBULA uniquely implements dual-axis evaluation (DAE) that separates capability assessment from stress testing, while all other benchmarks report only task-level success rates. Table \ref{table:nebulaVSbenchmark} also highlights NEBULA's comprehensive data collection: three modalities (RGB, depth, segmentation) and six camera viewpoints versus the single modality and 1-2 views standard in other benchmarks. Only LIBERO and VLABench match NEBULA's tiered difficulty structure, though neither provides the diagnostic isolation of specific capabilities that NEBULA's six distinct task families enable.